\pdfoutput=1

\documentclass[11pt]{article}

\usepackage[]{ACL2023}

\usepackage{times}
\usepackage{latexsym}

\usepackage[T1]{fontenc}

\usepackage[utf8]{inputenc}

\usepackage{microtype}

\usepackage{inconsolata}
\usepackage{array}

\usepackage{natbib}

\usepackage{graphicx}
\usepackage{subcaption}
\usepackage[normalem]{ulem}

\usepackage{tikz}
\usetikzlibrary{arrows}
\usepackage{tikz}

\usepackage{tikzscale}
\usetikzlibrary{arrows.meta,
                calc,
                positioning,
                quotes,
                shapes.geometric}

\pgfdeclareshape{datastore}{
  \inheritsavedanchors[from=rectangle]
  \inheritanchorborder[from=rectangle]
  \inheritanchor[from=rectangle]{center}
  \inheritanchor[from=rectangle]{base}
  \inheritanchor[from=rectangle]{north}
  \inheritanchor[from=rectangle]{north east}
  \inheritanchor[from=rectangle]{east}
  \inheritanchor[from=rectangle]{south east}
  \inheritanchor[from=rectangle]{south}
  \inheritanchor[from=rectangle]{south west}
  \inheritanchor[from=rectangle]{west}
  \inheritanchor[from=rectangle]{north west}
  \backgroundpath{
    \southwest \pgf@xa=\pgf@x \pgf@ya=\pgf@y
    \northeast \pgf@xb=\pgf@x \pgf@yb=\pgf@y
    \pgfpathmoveto{\pgfpoint{\pgf@xa}{\pgf@ya}}
    \pgfpathlineto{\pgfpoint{\pgf@xb}{\pgf@ya}}
    \pgfpathmoveto{\pgfpoint{\pgf@xa}{\pgf@yb}}
    \pgfpathlineto{\pgfpoint{\pgf@xb}{\pgf@yb}}
 }
}

\usepackage{adjustbox}
\usepackage{array} 

\newcolumntype{C}[1]{>{\centering\arraybackslash}m{#1}}

\makeatletter
\def\@fnsymbol#1{}
\makeatother

\title{

A Short Survey of Human Mobility Prediction in Epidemic Modeling from Transformers to LLMs

}

\author{
  {\bf \fontsize{10.5}{10.5}\selectfont Christian N. Mayemba$^{1}$} {\bf \fontsize{10.5}{10.5}\selectfont D'Jeff K. Nkashama$^{1,2}$ } {\bf \fontsize{10.5}{10.5}\selectfont Jean Marie Tshimula$^{1,3}$}
  {\bf \fontsize{10.5}{10.5}\selectfont Maximilien V. Dialufuma$^{1,7}$} \\  {\bf \fontsize{10.5}{10.5}\selectfont Jean Tshibangu Muabila$^{1,4}$} {\bf \fontsize{10.5}{10.5}\selectfont Mbuyi Mukendi Didier$^{1,3,8,9}$} 
  {\bf \fontsize{10.5}{10.5}\selectfont Hugues Kanda$^{1}$} {\bf \fontsize{10.5}{10.5}\selectfont René Manassé Galekwa$^{1,3,5}$} \\  {\bf \fontsize{10.5}{10.5}\selectfont Heber Dibwe Fita$^{1}$}  {\bf \fontsize{10.5}{10.5}\selectfont Serge Mundele$^{1}$} {\bf \fontsize{10.5}{10.5}\selectfont Kalonji Kalala$^{1,6}$} {\bf \fontsize{10.5}{10.5}\selectfont Aristarque Ilunga$^{1,3}$} {\bf \fontsize{10.5}{10.5}\selectfont \color{black}Lambert Mukendi Ntobo$^{10}$} \\ 
  {\bf \fontsize{10.5}{10.5}\selectfont \color{black} Dominique Muteba$^{11}$} {\bf \fontsize{10.5}{10.5}\selectfont Aaron Aruna Abedi$^{11}$} 
  \thanks{%
    $^{1}$Groupe de Recherche de Prospection et Valorisation des Données (Greprovad), Global $^{2}$GRIC, Université de Sherbrooke, Canada
    $^{3}$University of Kinshasa, Dem. Rep. of the Congo (DRC) $^{4}$LISV-UVSQ, Université Paris-Saclay, France $^{5}$University of Klagenfurt, Austria $^{6}$School of Electrical Engineering and Computer Science, University of Ottawa, Canada 
    $^{7}$Montreal Behavioural Medicine Centre, Centre Intégré Universitaire de Santé et Services Sociaux du Nord-de-l’Île-de-Montréal (CIUSSS-NIM), Canada $^{8}$Biomedical Research Unit, Hospital Monkole, Kinshasa, DRC $^{9}$University of Florida, USA $^{10}${\color{black}Programme National de Lutte Contre la Trypanosomiase Humaine Africaine (PNLTHA), Direction of Epidemiological Surveillance, Ministry of Public Health of the DRC} 
    $^{11}${\color{black}Centre National d'Intelligence Épidémiologique (CNIEP), Direction of Epidemiological Surveillance, Ministry of Public Health of the DRC}. Correspondence email: \href{mailto:jeanmarie.tshimula@unikin.ac.cd}{\texttt{jeanmarie.tshimula@unikin.ac.cd}} and \href{mailto:christian.mayemba@greprovad.org}{\texttt{christian.mayemba@greprovad.org}}
  }
}

\date{}

\begin{document}
\maketitle

\begin{abstract}


This paper provides a comprehensive survey of recent advancements in leveraging machine learning techniques, particularly Transformer models, for predicting human mobility patterns during epidemics. Understanding how people move during epidemics is essential for modeling the spread of diseases and devising effective response strategies. Forecasting population movement is crucial for informing epidemiological models and facilitating effective response planning in public health emergencies. Predicting mobility patterns can enable authorities to better anticipate the geographical and temporal spread of diseases, allocate resources more efficiently, and implement targeted interventions. We review a range of approaches utilizing both pretrained language models like BERT and Large Language Models (LLMs) tailored specifically for mobility prediction tasks. These models have demonstrated significant potential in capturing complex spatio-temporal dependencies and contextual patterns in textual data.

\end{abstract}

\section{Introduction}

Predicting population movements during disease outbreaks is a complex yet crucial task, with significant implications for public health decision-making and the formulation of epidemic control strategies. The recent COVID-19 pandemic has underscored the importance of understanding human mobility in predicting and controlling the spread of infectious diseases. Human mobility data can be combined with other data sources to help understand mobility patterns. This provides valuable insights into how to slow down the rapid spread of the disease. Additionally, it helps to analyze the correlation between the number of epidemic-infected cases and human activities in recreational areas such as parks. Furthermore, it enables early detection and prompt isolation of virus infection. Mobility data, derived from various sources such as call detail records, global positioning system, social networks, and expert knowledge of a region \cite{isaacman2012human,ebrahimpour2020analyzing, sobral2020ontology}.

While traditional epidemiological models heavily rely on mobility data, employing approaches like clustering techniques, differential equations, and statistical modeling \cite{kulkarni2019examining,rahman2021machine}, recent years have witnessed a paradigm shift towards the use of deep learning methodologies, specifically Transformer architectures pretrained on large corpora. These advanced techniques aim to tackle the inherent complexities involved in modeling human mobility dynamics during epidemics \cite{ma2022human,kobayashi2023modeling}.


\textbf{Contribution}. In this paper, we provide a comprehensive overview of recent research endeavors aimed at leveraging machine learning techniques, specifically Transformer models, to enhance the prediction of human mobility patterns in the context of epidemics. We highlight the contributions of both pretrained language models and Large Language Models (LLMs) tailored explicitly for mobility prediction tasks. Furthermore, we discuss the challenges and future directions in this emerging field, emphasizing the potential of these advanced modeling techniques to inform more accurate and actionable epidemiological models.

\begin{figure*}[t]
    \begin{minipage}[b]{0.5\textwidth}
        \centering
        \includegraphics[height=3cm]{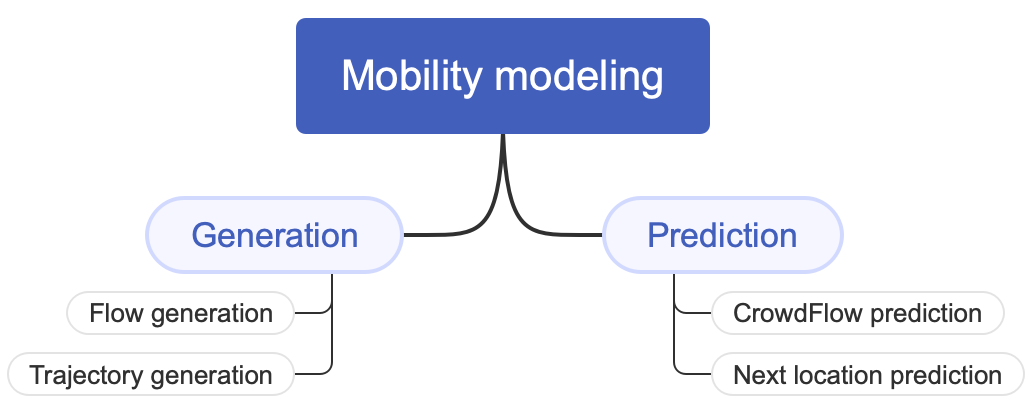}
        \caption{Human mobility modeling tasks taxonomy \\ by \citet{luca2021survey}.}
        \label{fig:human-mob}
    \end{minipage}
    \hfill
    \begin{minipage}[b]{0.5\textwidth}
        \centering
        \includegraphics[scale=0.5]{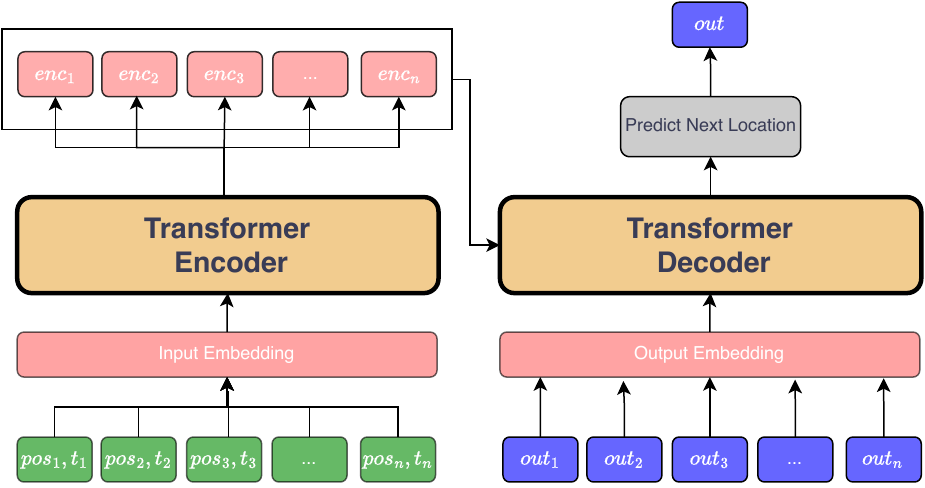}
        \caption{Transformer-based mobility prediction: Encoding mobility trajectories to predict future locations.}
        \label{fig:transformer_mobility}
    \end{minipage}
\end{figure*}
\section{Human Mobility Tasks}

A mobility pattern describes the movement of a considered population over a given observation period. Human mobility modeling tasks can be categorized into two main tasks: \textit{generation}, which involves generating realistic mobility data, and \textit{prediction}, which involves predicting future mobility patterns at both the individual and collective levels \cite{luca2021survey}. In the context of our study, a trajectory corresponds to a sequence of spatio-temporal information related to an individual's movement. Depending on the task at hand, trajectories can be aggregated by geographic regions. Given two regions, a flow represents the frequency of individuals moving inbound and outbound from one region ({\it the origin}) to another ({\it the destination}). Figure~\ref{fig:human-mob} displays, in one hand, the generation task, which includes subtasks such as flow generation and trajectory generation, where generative models, including Transformers and LLMs, play a crucial role. On the other hand, the prediction task includes crowd flow prediction and next location prediction, for which robust forecasting models are required.



Our paper presents a comprehensive overview of recent advancements in leveraging Transformers and LLMs for modeling human mobility patterns, particularly in the context of epidemic control (see Table~\ref{tab:my_label}). This table provides information about the methods, applications, and datasets utilized in recent studies involving Transformers and LLMs for modeling human mobility patterns in the context of epidemic modeling.

\section{Transformers in Human Mobility}



\begin{table*}[!h]
\fontsize{8.5}{8.5}\selectfont
\centering
\begin{tabular}{|C{5.0cm}|C{.5cm}|C{2.6cm}|C{2.5cm}|C{3.0cm}|}
\hline
\textbf{Paper} & \textbf{Year} & \textbf{Method} & \textbf{Application} & \textbf{Datasets} \\
\hline
WiFiMod: Transformer-based Indoor Human Mobility Modeling using Passive Sensing \cite{Trivedi_2021} & 2021 & WiFiMod (Transformer-based model) & Predicting indoor human mobility & Enterprise WiFi system logs \\
\hline
MobTCast: Leveraging Auxiliary Trajectory Forecasting for Human Mobility Prediction \cite{xue2021mobtcast} & 2021 & Transformer & Human mobility prediction & Gowalla, Foursquare-NYC
(FS-NYC), and Foursquare-Tokyo (FS-TKY) \\
\hline
Predicting Human Behavior with Transformer Considering the Mutual Relationship between Categories and Regions \cite{Osawa_2021} & 2021 & Transformer & Predicting human mobility & Not specified \\
\hline
TraceBERT—A Feasibility Study on Reconstructing Spatial–Temporal Gaps from Incomplete Motion Trajectories via BERT Training Process on Discrete Location Sequences \cite{Crivellari_2022} & 2022 & BERT & Trajectory reconstruction & Real-world large-scale trajectory dataset of short-term tourists~(CDRs) \\
\hline
Integrating Transformer and GCN for COVID-19 Forecasting \cite{Li_2022} & 2022 & Transformer and GCN & COVID-19 Forecasting & Nytimes Coronavirus (COVID-19) Data  \\
\hline
{ Large Language Models for Spatial Trajectory Patterns Mining \cite{Zhang_2023}} & 2023 & LLMs such as GPT-4 and Claude-2 & Anomaly detection in mobility data & GEOLIFE,  PATTERNS-OF-LIFE \\
\hline
How Do You Go Where? Improving Next Location Prediction by Learning Travel Mode Information using Transformers \cite{Hong_2022} & 2022 & Transformer & Next location prediction &  Green Class (GC) and Yumuv \\
\hline
GeoFormer: Predicting Human Mobility using Generative Pre-trained Transformer \cite{Solatorio_2023} & 2023 & GPT-based model & Predicting human mobility & HuMob Challenge 2023 datasets \\
\hline
Modeling and Generating Human Mobility Trajectories using Transformer with Day Encoding \cite{kobayashi2023modeling} & 2023 & Transformer with Day Encoding & Modeling and generating human mobility trajectories & HuMob dataset \\
\hline
CrowdFlowTransformer: Capturing Spatio-Temporal Dependence for Forecasting Human Mobility \cite{choya2023crowdflowtransformer} & 2023 & Transformer & Crowd flow forecasting & Not specified \\
\hline
TrafFormer: A Transformer Model for Predicting Long-term Traffic \cite{Tedjopurnomo_2023} & 2023 & Transformer & Long-term traffic prediction & METR-LA, PEMS-BAY \\
\hline
Where Would I Go Next? Large Language Models as Human Mobility Predictors \cite{Wang_2023} & 2023 & LMM & Human mobility prediction & GEOLIFE, FSQ-NYC \\
\hline
User Re-identification via Human Mobility Trajectories with Siamese Transformer Networks \cite{Wang_2023} & 2023 & Siamese Transformer network & User re-identification & Gowalla, Brightkite, and Foursquare (NYC, TKY) \\
\hline
{Exploring Large Language Models for Human Mobility Prediction under Public Events \cite{liang2023exploring}} & 2023 & LLM & Human mobility prediction under public events & Publicly available event information and taxi trip data \\
\hline
Learning Daily Human Mobility with a Transformer-Based Model \cite{wang2024learning} & 2024 & Transformer & Modelling human mobility & Tokyo Metropolitan Area \\
\hline
Health-LLM: Large Language Models for Health Prediction via Wearable Sensor Data \cite{kim2024health} & 2024 & LLM & Epidemic control & PM\-Data, LifeSnaps, GLOBEM, AW\_FB, MIT\-BIH, and MIMIC-III \\
\hline
Beyond Imitation: Generating Human Mobility from Context-aware Reasoning with Large Language Models \cite{shao2024beyond} & 2024 & LLM & Mobility generation & Tencent and Mobile Dataset \\
\hline
Large Language Models as Urban Residents: An LLM Agent Framework for Personal Mobility Generation \cite{Wang_2024a} & 2024 & LLM & Personal mobility generation & Not specified \\
\hline
MobilityGPT: Enhanced Human Mobility Modeling with a GPT model \cite{Haydari_2024} & 2024 & GPT & Mobility modeling & Real-world datasets \\
\hline
{COLA: Cross-city Mobility Transformer for Human Trajectory Simulation \cite{wang2024cola}} & 2024 & Transformer & Human trajectory simulation & GeoLife, Yahoo, New York, Singapore \\
\hline
\end{tabular}
\caption{Literature review of Transformers and LLMs for modeling human mobility patterns to epidemic control}
\label{tab:my_label}
\end{table*}

Transformers are a type of deep learning architecture that consists of two parts: an encoder and a decoder \cite{vaswani2017attention}. They have been instrumental in the recent breakthroughs we observe in various machine learning tasks. These include, but are not limited to, text-to-image generation, machine translation, and text summarization. One of the key factors contributing to the success of Transformers is the attention mechanism. This mechanism allows the model to prioritize the most relevant input data for tasks such as predicting the next word given a context. While its initial application was primarily on textual data, it has since been established that Transformers are effective across a multitude of applications, including forecasting, where they have shown superior performance compared to their predecessors \cite{vaswani2017attention,Trivedi_2021,Osawa_2021, Solatorio_2023,xu2023multimodal,kobayashi2023modeling,Tedjopurnomo_2023,Wang_2023,wang2024cola}. 

Furthermore, Transformers are multimodal, meaning they can combine data sources of different types, such as text, images, graphs, etc. Consequently, their use has seen a significant rise in recent years, including in the prediction of human mobility patterns for epidemic modeling \cite{li2021long,devyatkin2021approaches,xue2022translating,cui2021into,xue2021mobtcast,mai2022dengue,Li_2022,Hong_2022,shen2023forecasting,ren2023transcode,botz2022modeling,terashima2023human,bengio2020predicting,xu2021simulating,ma2022hierarchical,aragao2023covid,violos2022self,choya2023crowdflowtransformer,mao2023mpstan,wang2023pategail,chen2023mobcovid}. Figure~\ref{fig:transformer_mobility} illustrates the architecture of a Transformer model designed for mobility prediction. The model receives a sequence of location and time data representing a mobility trajectory as input. This trajectory is encoded using the Transformer's encoder component, which captures the temporal and spatial dependencies within the sequence. The resulting encoding is then passed to the decoder, which generates predictions for the next location in the trajectory. This self-contained framework utilizes the Transformer's attention mechanism to effectively capture long-range dependencies and spatial-temporal patterns in mobility data, enabling accurate prediction of future locations.

Initially, Transformer-based models like BERT showed promise in predicting mobility flows based on textual and location data \cite{devlin2018bert,li2021long,Crivellari_2022}. However, challenges persisted in generalization to new locations and outbreak scenarios \cite{devyatkin2021approaches}. \citet{terashima2023human} introduce LP-BERT for predicting human mobility trajectories using the Transformer architecture. LP-BERT enables parallel predictions, reducing training and prediction times, which can be beneficial for tasks like epidemic modeling that require quick insights into population movements.

When discussing epidemic modeling, \citet{botz2022modeling} discuss modeling approaches for early warning, monitoring of pandemics, and decision support in public health crises. It emphasizes the importance of population-level computational modeling, including machine learning techniques, in strengthening healthcare systems against respiratory infections. The authors highlight the significance of predicting outbreak impacts, monitoring disease spread, and assessing intervention effectiveness. 

Moreover, \citet{ma2022human} discuss the importance of human trajectory completion in controlling the spread of COVID-19, present a solution based on Transformers and evaluate it using an open-source human mobility dataset. The proposed solution involves using Transformers and deep learning models to estimate missing elements in trajectories.


Similarly, \citet{li2021long} present a Transformer-based model for long-term prediction of seasonal influenza outbreaks. The proposed model addresses the limitations of traditional forecasting methods by leveraging the Transformer's ability to capture long-range dependencies, and introduces a sources selection module based on curve similarity measurement to incorporate spatial dependencies.

To monitor human movements and comprehend the emergence of the pandemic, \citet{bengio2020predicting} develop advanced deep learning models for predicting infectiousness for proactive contact tracing during the COVID-19 pandemic, introduce the concept of proactive contact tracing (PCT) and discuss the use of deep learning predictors to locally predict individual infectiousness based on contact history while respecting privacy constraints. The study highlights the effectiveness of deep learning-based PCT methods in reducing disease spread compared to other tracing methods, suggesting their potential for deployment in smartphone apps to balance virus spread and economic costs while maintaining strong privacy measures.

Recent studies have made significant strides in leveraging advanced deep learning techniques for forecasting and modeling various aspects of the COVID-19 pandemic \cite{devyatkin2021approaches,cui2021into,violos2022self,xu2021simulating}. These studies utilize recurrent neural networks and Transformer-like architectures, multi-range encoder-decoder frameworks, self-attention based models, and generative adversarial networks to analyze socioeconomic impacts, forecast COVID-19 cases, predict human density in urban areas, and simulate human mobility trajectories.

\citet{devyatkin2021approaches} develop deep neural network models for forecasting the socioeconomic impacts of COVID-19 in Russian regions, particularly focusing on the regional cluster of Moscow and its neighbors. The models, based on recurrent and Transformer-like architectures, utilize heterogeneous data sources including daily cases, age demographics, transport availability, and hospital capacity. The study shows that incorporating demographic and healthcare features improves the accuracy of economic impact predictions, and data from neighboring regions enhances predictions of healthcare and economic impacts. Overall, the research emphasizes the importance of forecasting to address inter-territorial inequality during the pandemic. \citet{cui2021into} propose a multi-range encoder-decoder framework for COVID-19 prediction, leveraging historical case data, human mobility patterns, and reported cases and deaths to enhance prediction accuracy. By embedding features from multiple expose-infection ranges and utilizing message passing between time slices, the model surpasses existing methods in both weekly and daily prediction tasks. Ablation studies confirm the effectiveness of key components, demonstrating the model's ability to perform well with or without mobility data. The framework addresses challenges posed by incomplete data and unknown disease factors, offering a promising approach for precise and timely COVID-19 forecasting.

\citet{violos2022self} present a self-attention based encoder-decoder model for predicting human density in urban areas, incorporating deep learning methods and geospatial feature preprocessing. This research enhanced human mobility prediction in epidemic modeling by providing insights into population movement patterns, aiding in the analysis of disease transmission dynamics, and supporting the implementation of strategic interventions to mitigate the spread of epidemics. \citet{xu2021simulating} propose DeltaGAN, a generative model for synthesizing continuous-time human mobility trajectories. DeltaGAN captures realistic mobility dynamics without discretizing visitation times, enabling more accurate trajectory generation and analysis.  Its utility is demonstrated in studying the spreading of COVID-19, showing small divergence in population distribution compared to real data.

Spatio-temporal epidemic forecasting models have been developed to predict epidemic transmission dynamics by integrating domain knowledge with neural networks \cite{mao2023mpstan,ma2022hierarchical}. \citet{mao2023mpstan} introduce a spatio-temporal epidemic forecasting model called MPSTAN, which integrates domain knowledge with neural networks to accurately predict epidemic transmission. This study emphasizes the importance of selecting appropriate domain knowledge for forecasting and proposes a dynamic graph structure to capture evolving interactions between patches over time. \citet{ma2022hierarchical} introduce an approach, Hierarchical Spatio-Temporal Graph Neural Networks (HiSTGNN), for pandemic forecasting using large-scale mobility data. HiSTGNN incorporates a two-level neural architecture and a Transformer-based model to capture spatial and temporal information hierarchically. The model outperforms existing baselines in predicting COVID-19 case counts, demonstrating its superior predictive power. The research highlights the importance of leveraging mobility data for pandemic forecasting and addresses the limitations of existing Graph Neural Networks in capturing community structures within mobility graphs.

Additionally, models like CF-Transformer and MSP-STTN have been proposed to capture spatio-temporal dependencies for crowd flow forecasting, contributing to human mobility prediction in epidemic modeling \cite{choya2023crowdflowtransformer,xie2022multisize}. More specifically, \citet{choya2023crowdflowtransformer} introduced the CrowdFlowTransformer (CF-Transformer) model, which combines Transformer with graph convolution to capture spatio-temporal dependencies for crowd flow forecasting, and aims to improve forecasting accuracy by considering both temporal and spatial aspects of crowd flow data for applications like human mobility prediction in epidemic. \citet{xie2022multisize} proposed the MSP-STTN model for short- and long-term crowd flow prediction, focusing on grid-based crowd data analysis. MSP-STTN  contributes to human mobility prediction in epidemic modeling by providing insights into long-term crowd flow patterns, aiding in urban planning and traffic management. Its applications extend to various grid-based prediction problems beyond crowd flow analysis, such as weather forecasting and air pollution prediction.

These advancements underscore the critical role of machine learning in enhancing our understanding of disease dynamics and informing public health interventions during epidemics.

\section{Large Language Models in Human Mobility}

Recently, there has been a surge in the development of Large Language Models (LLMs) tailored specifically for high-fidelity human mobility simulation and forecasting \cite{xue2022leveraging,liang2023exploring,wang2023would,Zhang_2023,tang2024synergizing,shao2024beyond,kim2024health,wang2024cola,Haydari_2024}. These models, trained on massive corpora of mobility data paired with auxiliary information, demonstrate the capability to generate plausible mobility trajectories for entire populations under various policy and disease conditions. Despite these advancements, challenges persist regarding ensuring adequate coverage, transparency, and safety for real-world epidemiological applications.

Exploring further, \citet{xue2022leveraging} propose a pipeline that leverages language foundation models for human mobility forecasting by transforming numerical temporal sequences into sentences for prediction tasks. By integrating language models with mobility prompts, the study provides empirical evidence of the effectiveness of this approach in discovering sequential patterns, which can be valuable for predicting human mobility in epidemic modeling scenarios and potential disease spread. Similarly, \citet{liang2023exploring} explore LLMs' application for predicting human mobility patterns during public events (LLM-MPE). Addressing the challenge of incorporating textual data from online event descriptions into mobility prediction models, LLM-MPE transforms raw event descriptions into a standardized format and segments historical mobility data to make demand predictions considering both regular and event-related components. This approach can indirectly inform epidemic modeling by providing insights on travel patterns and potential disease spread dynamics during events, thereby aiding the development of more accurate epidemic models.


In another stride, \citet{wang2023would} introduce LLM-Mob, a framework utilizing LLMs for human mobility prediction, capturing both long-term and short-term dependencies and employing context-inclusive prompts. LLM-Mob contributes to epidemic modeling by providing interpretable predictions, underscoring the potential of LLMs in advancing human mobility prediction techniques to address epidemic spread.

\citet{tang2024synergizing} present an approach that integrates LLMs with spatial optimization for urban travel itinerary planning. Focusing on the Online Urban Itinerary Planning (OUIP) problem, this study demonstrates the effectiveness of the proposed system through offline and online experiments. The methodology involves using LLMs like GPT-3.5 and GPT-4 for itinerary generation, along with spatial optimization techniques and rule-based metrics for evaluation. This approach can contribute to human mobility prediction in epidemic modeling by efficiently generating personalized and coherent itineraries based on natural language requests, which can help understand and predict human movement patterns in urban contexts during epidemics. Furthermore, by leveraging LLMs for itinerary generation and spatial optimization, the system can adapt to diverse user needs and provide tailored travel plans, valuable in modeling and predicting human mobility changes during epidemics for better public health planning and management.

More recently, \citet{shao2024beyond} proposed an approach called MobiGeaR for generating human mobility data using LLMs and a mechanistic gravity model. MobiGeaR involves leveraging LLM reasoning and a divide-and-coordinate mechanism to generate mobility patterns effectively. The proposed approach significantly reduces the token cost per trajectory and boosts the accuracy of mobility prediction models through data augmentation. The MobiGeaR approach can contribute to human mobility prediction in epidemic modeling by generating high-quality data to augment sparse datasets, enabling mining and modeling of motion patterns for predicting future trajectories based on historical data. The approach can improve the predictive performance crucial for epidemic control and other applications requiring accurate mobility by providing better enhancements in downstream mobility prediction tasks, particularly in intention-type prediction.

\section{Challenges and Limitations}

Despite their promising performance, Transformers and LLMs face several challenges when applied to human mobility prediction tasks in epidemic modeling. One major challenge is the availability and quality of relevant data sources, which can be subject to biases or errors that affect model performance \cite{kulkarni2019examining}. Additionally, the applicability of these advanced models extends beyond well-resourced regions to low- and middle-income countries (LMICs) and resource-constrained settings with underdeveloped electronic health records \cite{tshimula2023redesigning}. In these contexts, leveraging machine learning techniques for human mobility prediction can significantly enhance the understanding and management of epidemics by providing valuable insights into population movements even with limited data availability and infrastructure.

{\color{black}
Mastering the speed of mobility and the number of movements within a given environment during an epidemic context can consequently help in formulating appropriate public health strategies. Taking the example of a screening activity for sleeping sickness in a village where the endemic level is known, and where the main activity of the inhabitants is farming, with mobility defined between the place of residence and the fields during dawn and dusk hours, the failure to consider this type of mobility by healthcare professionals could result in a large number of absences and non-respondents to these activities, even though these individuals had been planned and accounted for.

On the other hand, considering an industrialized country context, where means of transportation include airplanes, subways, high-speed trains, and where large surfaces and amusement parks are present, the speed and number of movements would also be high; in such an environment, the spread rate of an epidemic would be directly proportional to mobility. It is therefore important to master the mobility data of such a population and to use it in a public health context to contain the epidemic.}

Implementing artificial intelligence (AI) models in LMICs poses significant challenges, primarily due to the potential non-reproducibility of their initial performance upon integration with local datasets and the absence of regulatory frameworks \cite{wang2023chatgpt}. Addressing this challenge is critical to ensuring the effectiveness and reliability of Transformers or LLMs used for human mobility modeling in LMICs, ultimately enhancing epidemiological surveillance and the health outcomes of local populations. While fine-tuning these AI models is a recommended approach for specific applications \cite{yang2023harnessing,li2019fine}, it is essential, particularly in the context of LMICs, to plan cross-validation of these models with local datasets to improve and reproduce the model's original performance.

Moreover, ethical considerations may arise when using these models for surveillance purposes or making decisions about public health interventions based on predictions from these models. Therefore, ensuring responsible deployment of these technologies, particularly in underserved regions, is crucial for achieving equitable and effective epidemic control strategies.

%

\section{Conclusion}

This emerging area shows promise for improving epidemiological modeling through advanced mobility prediction. Continued progress in integrating multimodal data streams and expert knowledge can significantly bolster public health decision-making by providing more realistic models of human movement dynamics during crises. However, further work is essential to overcome existing limitations and ensure responsible deployment of LLMs.

The successful implementation of Transformers or LLMs models in LMICs necessitates careful consideration of the model's suitability for the local context and adjustments to the training and validation datasets. The scope of implementing these AI models in LMICs lies in developing more contextually appropriate models, integrating local datasets, and fostering collaboration to improve performance and reproducibility.

Future research endeavors should prioritize enhancing model generalizability across diverse geographical and socio-economic contexts. Moreover, efforts should be directed towards adapting these advanced modeling techniques to resource-constrained settings, particularly in LMICs, where access to data and computational resources may be limited. This includes exploring innovative approaches for collecting and processing human mobility data in LMICs, as well as adapting LLMs to accommodate varying sociocultural contexts.

Addressing these challenges will be crucial for ensuring the widespread applicability and impact of machine learning-based approaches in epidemic modeling and public health decision-making worldwide. This will contribute to more equitable and effective epidemic response strategies on a global scale.

\section*{Acknowledgments}
The authors thank all Greprovad members for helpful discussions and comments on early drafts.

\bibliography{custom.bib}
\bibliographystyle{acl_natbib.bst}

\end{document}